\documentclass{article}
\usepackage{spconf,amsmath,graphicx}

\usepackage{enumitem}
\setlist{nosep, leftmargin=14pt}

\usepackage{mwe} % to get dummy images
\usepackage{adjustbox}
\usepackage{xcolor}
\usepackage{booktabs} 
\usepackage{xcolor}

\usepackage{amsfonts}
\usepackage[skip=2pt]{caption} 
\definecolor{lightblue}{RGB}{16,98,180}
\definecolor{lightpink}{RGB}{243,40,109}
\definecolor{lightgreen}{RGB}{0,220,0}

\usepackage{subcaption}

\title{Learning to Reason About Rare Diseases \\through Retrieval-Augmented Agents}

\name{\parbox{\textwidth}{Ha Young Kim\textsuperscript{1, 2}, Jun Li\textsuperscript{1, 3}, Ana Beatriz Solana\textsuperscript{2}, Carolin M. Pirkl\textsuperscript{2}, Benedikt Wiestler\textsuperscript{1, 3, 4}, \\Julia A. Schnabel\textsuperscript{1, 3, 5, 6, *}, Cosmin I. Bercea\textsuperscript{1, 3, *}\thanks{* Co-Senior authors}, On behalf of the PREDICTOM consortium }}

\address{%
  \begin{minipage}{\linewidth}\small
    \centering
    \textsuperscript{1}Technical University of Munich, Munich, Germany \\ \textsuperscript{2}GE HealthCare, Munich, Germany\\ 
    \textsuperscript{3}Munich Center for Machine Learning, Munich, Germany\\
    \textsuperscript{4}Klinikum Rechts der Isar, Munich, Germany\\
    \textsuperscript{5}Helmholtz Munich, Munich, Germany\\  \textsuperscript{6}King's College London, London, United Kingdom\\  
  \end{minipage}%
}

\begin{document}
%\ninept
%
\maketitle
\begin{abstract}
Rare diseases represent the long tail of medical imaging, where AI models often fail due to the scarcity of representative training data. In clinical workflows, radiologists frequently consult case reports and literature when confronted with unfamiliar findings. Following this line of reasoning, we introduce \textit{RADAR} (\textit{Retrieval-Augmented Diagnostic Reasoning Agents}), an agentic system for rare disease detection in brain MRI. Our approach uses AI agents with access to external medical knowledge by embedding both case reports and literature using sentence transformers and indexing them with FAISS to enable efficient similarity search. The agent retrieves clinically relevant evidence to guide diagnostic decision-making on unseen diseases, without the need of additional training. Designed as a model-agnostic reasoning module, RADAR can be seamlessly integrated with diverse large–language models, consistently improving their rare pathology recognition and interpretability. On the NOVA dataset comprising 280 distinct rare diseases, RADAR achieves up to a 10.2\% performance gain, with the strongest improvements observed for open-source models such as DeepSeek. Beyond accuracy, the retrieved examples provide interpretable, literature-grounded explanations, highlighting retrieval-augmented reasoning as a powerful paradigm for low-prevalence conditions in medical imaging. Code and Details: https://anonymous.4open.science/r/RADAR-3232
\end{abstract}
\section{Introduction}

Rare brain disorders collectively affect millions worldwide, yet their diagnosis remains notoriously difficult due to the scarcity of expert knowledge, the heterogeneity of clinical presentations, and the extremely low prevalence of individual conditions. These challenges often result in misdiagnosis, delayed interventions, and inappropriate treatments, underscoring the need for diagnostic systems that are not only accurate but also interpretable and evidence-grounded \cite{schieppati2008rare}.

\label{sec:intro}
\begin{figure}[t]
  \centering
  \centerline{\includegraphics[width=8.5cm]{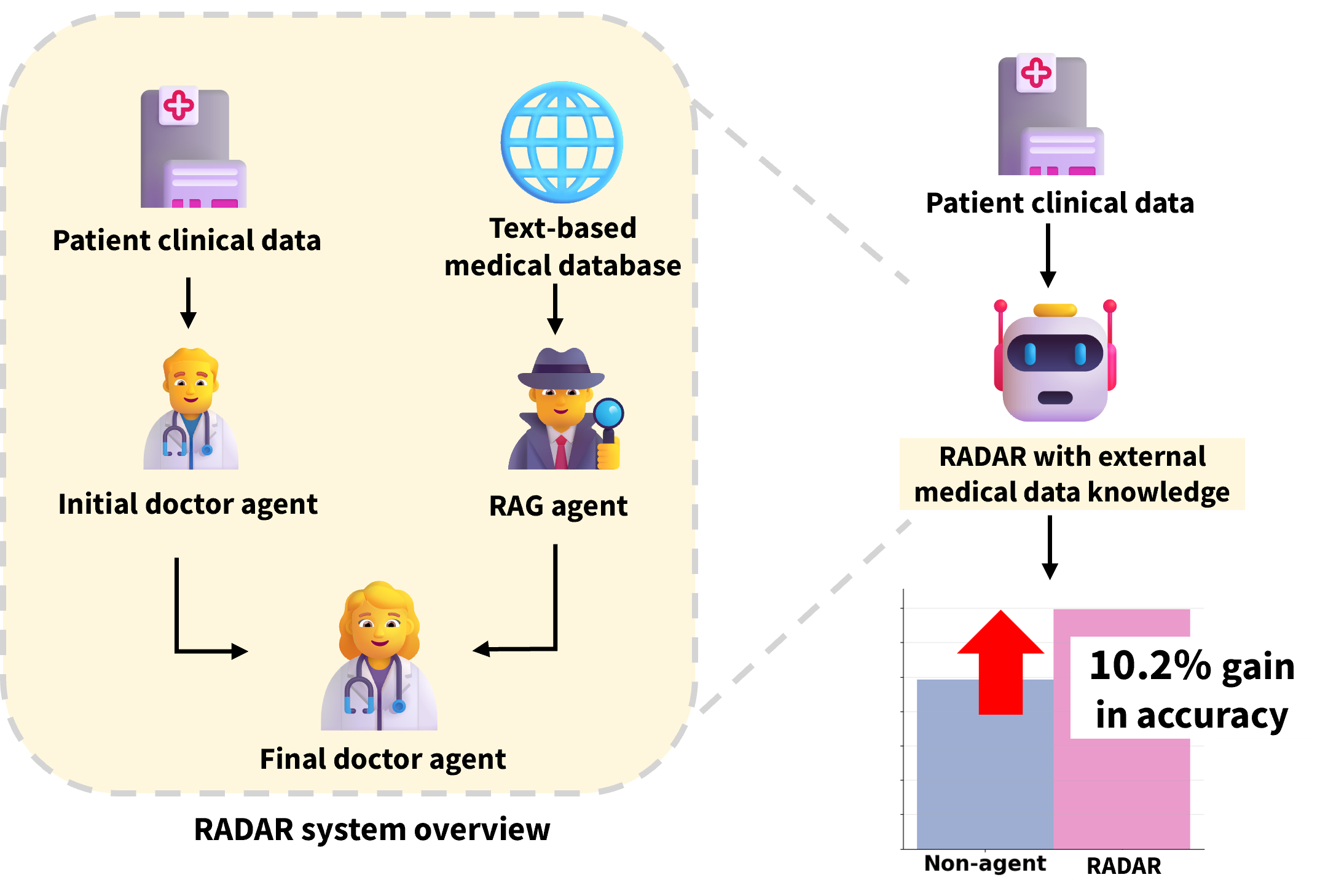}}
\caption{Overview of the proposed RADAR (Retrieval-Augmented Diagnostic Reasoning Agents) framework. The system employs coordinated agents that retrieve and integrate medical knowledge from external text-based databases to support diagnostic reasoning on rare diseases, achieving up to a 10.2\% accuracy gain over non-agentic baselines.}
\label{fig:first}
\end{figure}

Recent advances in artificial intelligence (AI), especially large language models (LLMs) and agentic AI systems, have demonstrated high potential in complex reasoning and clinical decision support across medical domains~\cite{llmclinicalreasoner, savage2024diagnostic}. However, the current high-performing LLM-based systems operate as closed models and lack domain-specific medical training. Consequently, this often leads to misdiagnoses or hallucinated recommendations in clinical settings \cite{zhu-etal-2025-trust}.

Retrieval-augmented generation (RAG) addresses these limitations by combining the generative reasoning capabilities of LLMs with real-time access to sources of domain-specific knowledge \cite{ragreview, gargari2025rag}. This paradigm has a notable resemblance to clinical workflows: when radiologists encounter unfamiliar findings, they consult case reports and literature to guide their reasoning. Inspired by this clinical workflow, we propose RADAR (Retrieval-Augmented Diagnostic Reasoning Agents), an agentic system for retrieval-augmented diagnostic reasoning that can dynamically retrieve relevant medical evidence during inference, without requiring additional fine-tuning or domain-specific retraining. This approach enhances diagnostic accuracy while simultaneously mitigating hallucinations and ensuring evidence-based outputs. By explicitly linking model decisions to supporting evidence, RADAR reduces hallucinations and enables transparent, literature-backed decisions.

RADAR employs a set of coordinated agents that (i) retrieve semantically relevant case reports and literature from external sources, (ii) ground diagnostic reasoning on retrieved evidence, and (iii) synthesize interpretable diagnostic hypotheses. Our main contributions are:

\begin{itemize}
    \item We introduce RADAR, a retrieval-augmented, model-agnostic reasoning framework that integrates radiological understanding from brain MRI with external medical knowledge, improving diagnostic accuracy and interpretability without additional training.
    \item We conduct a comprehensive evaluation on the NOVA dataset covering 280 rare brain diseases, demonstrating that retrieval-based reasoning consistently improves both diagnostic accuracy and interpretability across diverse vision–language models, establishing a scalable path toward trustworthy AI in data-scarce medical imaging.
\end{itemize}

\section{Method}
\label{sec:method}

We propose {RADAR}(Retrieval-Augmented Diagnostic Reasoning Agents), a system designed to improve diagnostic reasoning for rare diseases by integrating multi-agent collaboration with retrieval-augmented generation as illustrated in Figure~\ref{fig:agent_sys} (d). RADAR iteratively generates diagnostic hypotheses, retrieves external medical knowledge, and refines diagnosic conclusions. The framework comprises three specialized agents: an \textit{initial doctor agent}, a \textit{retrieval agent}, and a \textit{final doctor agent}, that collaborate to produce an interpretable, evidence-grounded diagnosis.

\subsection{Initial Doctor Agent}
\label{sec:init_doc}
The initial doctor agent is implemented as a large language model (LLM) prompted to act as a diagnostic expert. It takes the MRI image findings (image caption) and patient clinical history data as input and produces a list of ten candidate diagnoses:
\[
f_{\text{init}}: C = (\text{Image caption}, \text{Clinical data}) \mapsto \{d_i\}_{i=1}^{10}.
\]
To promote diagnostic diversity, the underlying LLM is configured with a high temperature and top-$p$ sampling. This stage provides a broad but plausible hypothesis to guide further reasoning.

% \paragraph{(3) Answer Generation.}
% For each question \( q_i \), the agent retrieves the top-$k$ most relevant chunks based on cosine similarity:
% \[
% R_i = \text{Top-}k(\mathbf{v}_{q_i}, \mathcal{I}, k=5),
% \]
% where \(\mathbf{v}_{q_i} = g_{\text{embed}}(q_i)\).
% The LLM then summarizes the retrieved content to generate a concise, evidence-grounded answer. A low temperature is used to ensure factual precision over creativity.

\subsection{RAG agent}
The RAG agent assists the diagnostic reasoning process by generating targeted queries, retrieving external evidence, and synthesizing contextual answers. It operates in three stages:\\

\noindent \textbf{1. Query generation.} The query generation system takes the image caption and patient clinical data as input and uses an LLM to generate a set of question-keyword pairs: $$f_{\text{LLM}}: C = (\text{Image caption}, \text{Clinical data})\mapsto\{(q_i, k_i)\}_{i=1}^n$$ where $q_i$ is a question designed to enhance diagnostic reasoning, and $k_i$ is the corresponding keyword extracted from $q_i$. To generate a wide variety of search queries helpful for diagnosis, the LLM is set with a high temperature and top-$p$. \\

\noindent \textbf{2. Knowledge retrieval and indexing.} The knowledge retrieval and indexing system retrieves relevant information for each keyword $k_i$ as follows:
\begin{itemize}
\item Internal check: If information about the keyword $k_i$ exists in the internal knowledge base, it retrieves relevant documents. Otherwise, the system updates its knowledge base dynamically by constructing a new index through the retrieval of new data from an external source.

\item External retrieval: The agent queries Radiopaedia \cite{radiopaedia} for each keyword $k_i$, retrieving 10 relevant documents, 5 from the articles section and 5 from the cases section.
Each retrieved document is segmented into overlapping chunks \(\{c_{i,j}\}_{j=1}^{M_i}\).
These are embedded into dense vector embeddings using using the all-MiniLM-L6-v2 sentence transformer model ($g_{\text{embed}}$) \cite{all-minilm-l6-v2-model}.
\[
\mathbf{v}_{i,j} = g_{\mathrm{embed}}(c_{i,j}), \quad \mathbf{v}_{i,j} \in \mathbb{R}^{d}.
\]

Then, these embeddings are stored in the internal knowledge base as a FAISS index \cite{douze2025faisslibrary} to enable efficient similarity search, i.e., cosine similarity:
\[
\mathcal{I} \leftarrow \mathcal{I} \cup \{\mathbf{v}_{i,j}\}_{j=1}^{M_i}.
\]
\\
\end{itemize}

\noindent \textbf{3. Answer generation.} For each question \( q_i \), the agent retrieves the top-$k$ most relevant chunks based on cosine similarity:
\[
R_i = \text{Top-}k(\mathbf{v}_{q_i}, \mathcal{I}, k=5),
\]
where \(\mathbf{v}_{q_i} = g_{\text{embed}}(q_i)\). Using this, an LLM analyzes the retrieved content and generates a concise answer to the question $q_i$ based on the retrieved evidence. We configure this LLM with a low temperature to generate an answer based solely on the retrieved content.

\begin{figure*}[t]
\begin{subfigure}[t]{.15\linewidth}
  \centering
  \centerline{\includegraphics[height=5.5cm]{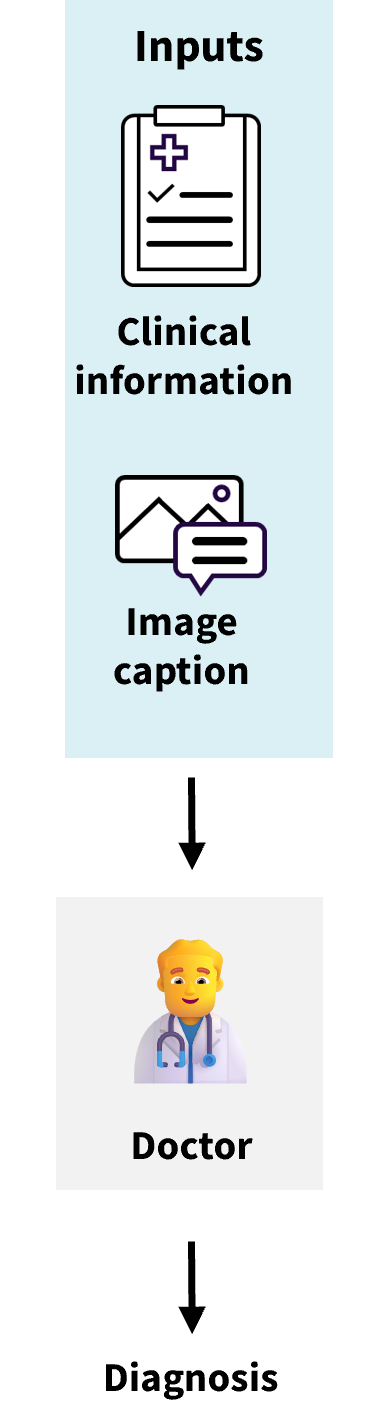}}
  \centerline{(a) Single agent}\medskip
\end{subfigure}
\begin{subfigure}[t]{.20\linewidth}
  \centering
  \centerline{\includegraphics[height=5.5cm]{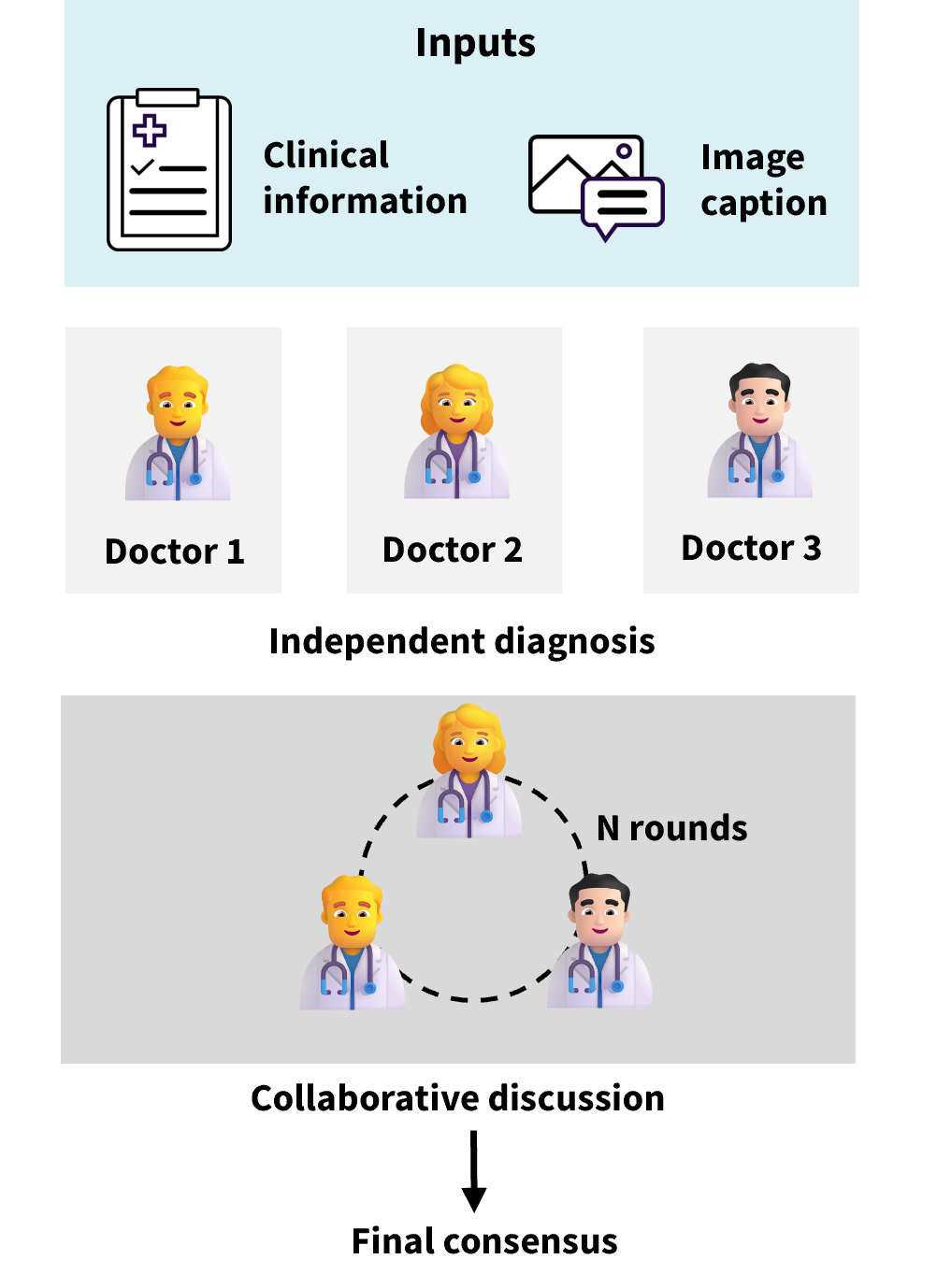}}
  \centerline{(b) Collaborative system}\medskip
\end{subfigure}
\begin{subfigure}[t]{.20\linewidth}
  \centering
  \centerline{\includegraphics[height=5.5cm]{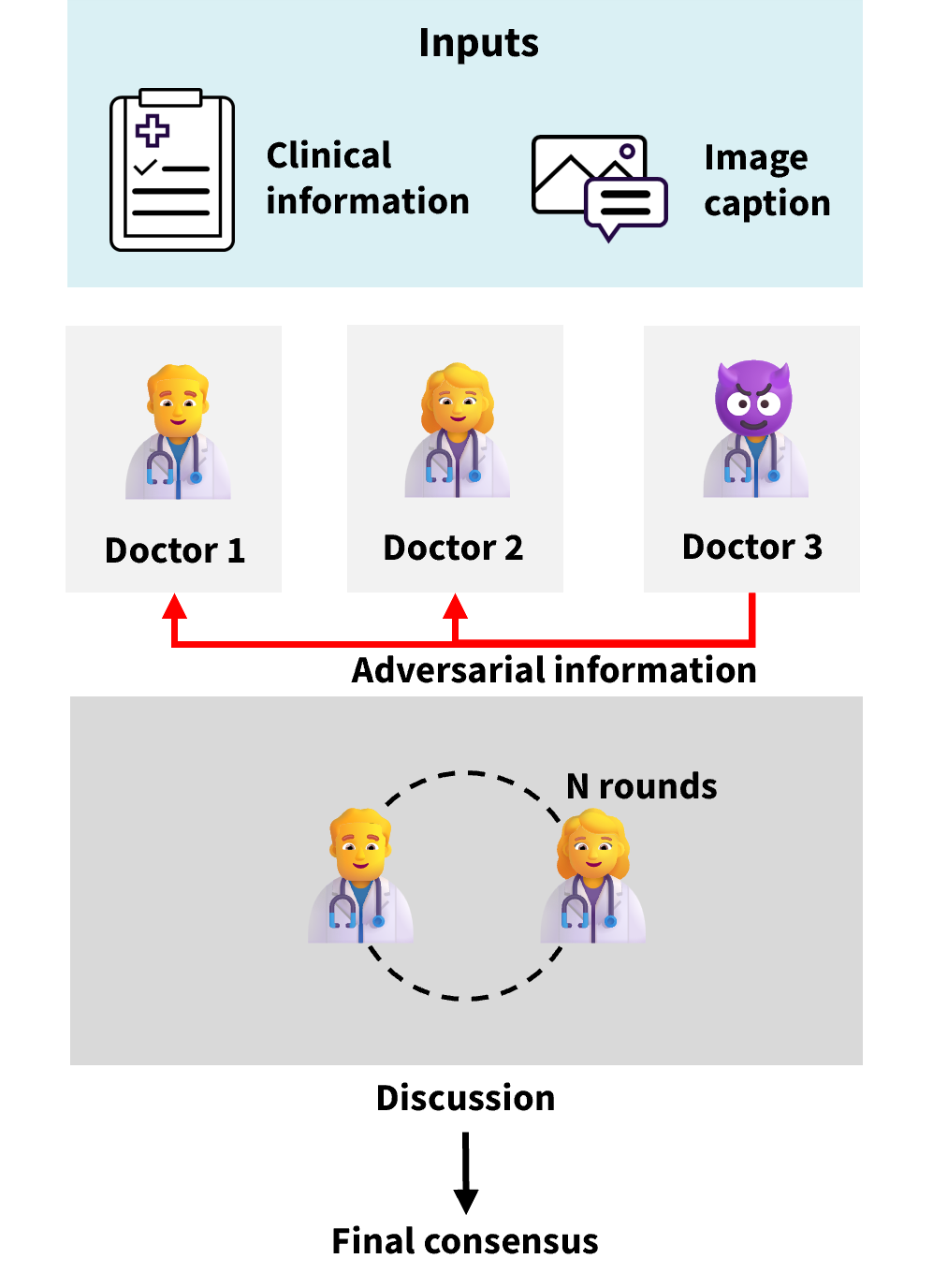}}
  \centerline{(c) Challenger system}\medskip
\end{subfigure}
\begin{subfigure}[t]{.45\linewidth}
  \centering
  \centerline{\includegraphics[width=7.0cm]{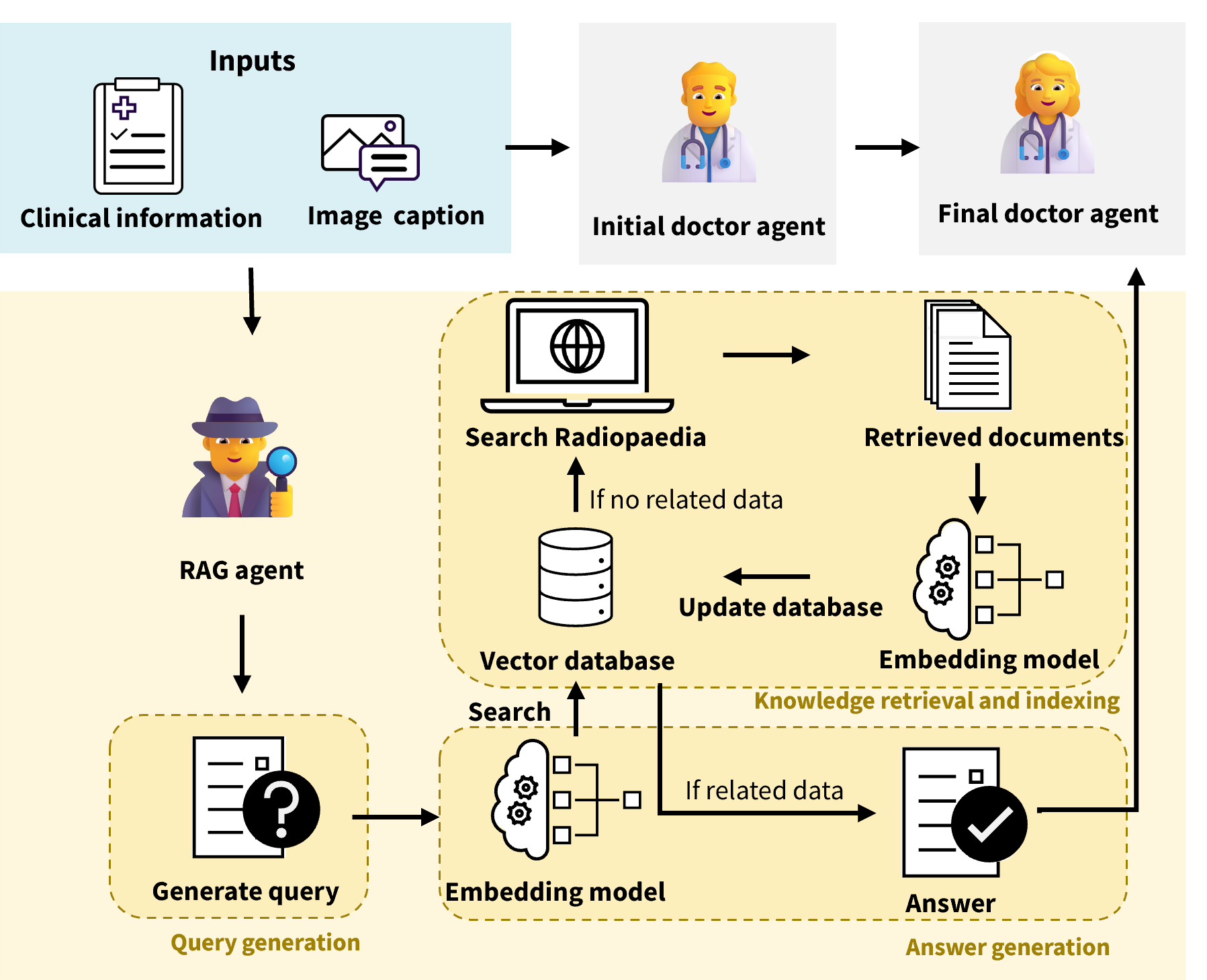}}
  \centerline{(d) RADAR system (ours)}\medskip
\end{subfigure}
\caption{Comparison of multi-agent diagnostic reasoning setups. (a) A single-agent system: a single doctor agent generates a diagnosis. (b) Collaborative system: agents exchange independent diagnoses and reach a consensus through discussion rounds. (c) Challenger system: one agent introduces adversarial information to test the robustness of others, and (d) RADAR (ours): retrieval-augmented framework where agents access external medical knowledge via Radiopaedia to refine and ground diagnostic reasoning. }
\label{fig:agent_sys}
\end{figure*}

\subsection{Final doctor agent}
% \noindent  \textbf{Final doctor agent} 
The final doctor agent integrates all the available data including image captions, clinical information, retrieved knowledge, and the candidate diagnosis, to produce one primary diagnosis and four differential diagnoses: 
\[
f_{\text{final}}: (C, \{d_i\}, R) \mapsto D_{\text{final}} = \{d_{\text{primary}}, d_{\text{diff}}^{(1-4)}\}.
\]
This agent operates at a mid-range temperature setting to balance reasoning flexibility with factual consistency. By explicitly conditioning on retrieved evidence, it produces interpretable, literature-grounded diagnostic outputs.

\section{Experiments}\label{sec:experiments}
Our experiments aim to assess three key aspects of {RADAR}:  
(1) whether retrieval-augmented reasoning improves diagnostic accuracy for rare brain diseases,  
(2) whether the system generalizes across diverse large language models (LLMs), and  
(3) whether the retrieved evidence enhances interpretability and trustworthiness.  

\subsection{Dataset and Metrics}
We evaluate RADAR on the publicly available {NOVA dataset} \cite{bercea2025nova}, which includes around 900 brain MRI scans spanning 281 rare pathologies and multiple acquisition protocols. Each case provides patient clinical information and an expert-written caption describing the imaging findings.
To mitigate potential bias toward single phrasing we paraphrased each caption into four alternative formulations using GPT-4o.

We measure the diagnostic performance using {Top-1} and {Top-5 accuracy}, following the evaluation protocol described in the original NOVA paper \cite{bercea2025nova}. Top-1 indicates exact agreement with the ground-truth diagnosis, while Top-5 considers whether the correct diagnosis appears among the five most likely predictions. Because medical terminology may differ across sources, each prediction is normalized via GPT-4o to align synonyms and variant expressions.

\subsection{Baselines and Model Configurations}
To ensure a fair and comprehensive comparison, we evaluate multiple reasoning paradigms across both closed- and open-source LLMs.  
Specifically, we test two proprietary models (GPT-4o \cite{achiam2023gpt4} and Gemini-2.0-Flash \cite{gemini2024}) and three open-source models (Qwen3-32B \cite{qwen3}, DeepSeek-R1-70B \cite{deepseek}, and MedGemma-27B \cite{medgemma}).  
% Each model is assessed under three reasoning setups: a single-agent configuration, a collaborative multi-agent system where doctors iteratively reach consensus (Figure~\ref{fig:agent_sys}a), and a challenger system introducing adversarial reasoning between agents (Figure~\ref{fig:agent_sys}b).  
% RADAR extends these setups by integrating retrieval-augmented reasoning, allowing agents to access external medical knowledge to ground and refine their diagnostic decisions as shown in Figure~\ref{fig:agent_sys}c.  
% Implementation details and prompts are provided in the repository.

Each model is evaluated under three reasoning setups: a single-agent system (Figure~\ref{fig:agent_sys}a), a collaborative multi-agent system (Figure~\ref{fig:agent_sys}b), and a challenger multi-agent system (Figure~\ref{fig:agent_sys}c). In the \textit{single-agent system} an LLM generates a diagnostic output directly from the patient information. The \textit{collaborative system} involves three independent doctor agents who provide initial diagnoses and subsequently engage in iterative discussions to reach consensus. In contrast, the \textit{challenger system} introduces an adversarial agent that challenges the reasoning of the doctor agents. \textit{RADAR} extends these setups by integrating retrieval-augmented reasoning, allowing agents to access external medical knowledge to ground and refine their diagnostic decisions as shown in Figure~\ref{fig:agent_sys}d. 
Implementation details and prompts are available in the repository.

\begin{figure*}[t]
  \centering
  \centerline{\includegraphics[width=16cm]{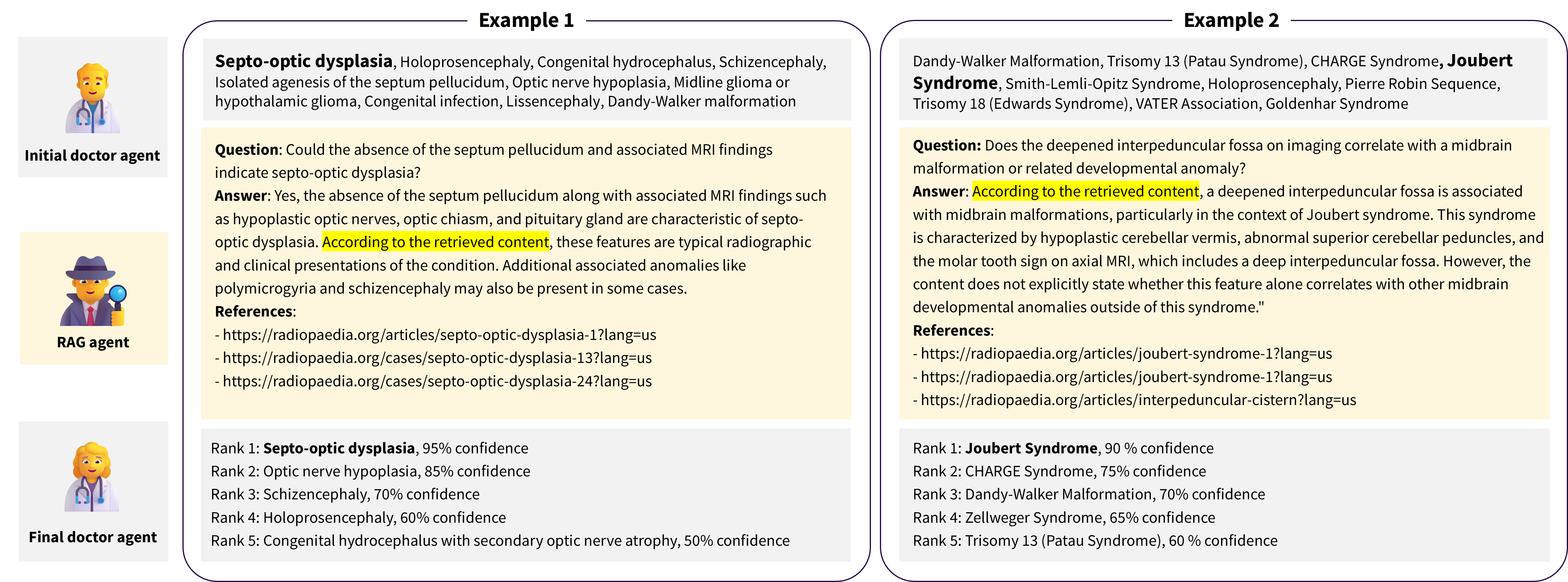}}
\caption{Examples of the results generated by our RADAR system. The ground-truth diagnosis is marked with bold.}

\label{fig:example_out}
\end{figure*}

\section{Results and discussion}
\label{sec:result}

\subsection{Results}
We evaluate RADAR against the single-agent, collaborative, and challenger multi-agent configurations. Tables~\ref{table:top1acc} and~\ref{table:top5acc} summarize Top-1 and Top-5 diagnostic accuracy across the five LLM backbones.
RADAR consistently outperforms all baselines across models and metrics.  
For instance, it achieves up to \textbf{+7.97} Top-1 improvement with Qwen3-32B and \textbf{+10.19} Top-5 improvement with DeepSeek-R1-70B over the single-agent baseline. Performance gains are especially pronounced for open-source models, suggesting that retrieval-augmented reasoning compensates for smaller model capacity and limited medical pretraining. 
The collaborative and challenger setups provide limited or inconsistent gains over the single-agent baseline, and in some cases, even degrade performance. This suggests that multi-agent interaction can amplify diagnostic uncertainty, particularly when the agents lack domain-specific knowledge. Our best performance was achieved using GPT-4o on our RADAR method, reaching Top-1 accuracy of 54.40\% and Top-5 accuracy of 75.05\%. For context, the NOVA paper~\cite{bercea2025nova} reports resident neuroradiologist performance of 48–52\% Top-1 and 68–76\% Top-5 accuracy, measured on a 25-case subset. RADAR achieves comparable accuracy evaluated across the full dataset, underscoring its potential of diagnostic reasoning agentic system in clinical settings. 
Nevertheless, RADAR still relies on radiologist-provided captions. Bridging this gap from textual reasoning to direct visual understanding remains an open challenge. 

Figure~\ref{fig:example_out} shows illustrative outputs from RADAR.  
It demonstrates how the RAG agent formulates targeted diagnostic questions, retrieves relevant content from Radiopaedia, and generates concise evidence-based answers strictly from the retrieved text. The final doctor agent integrates this information and gives a ranked list of differential diagnoses with confidence estimates.  
In the second example, retrieved evidence causes RADAR to update its prediction, recovering the correct diagnosis that was initially ranked lower. This shows the system’s capacity to adjust its reasoning by integrating new clinical information.

% An example output from our RADAR system is presented in Figure~\ref{fig:example_out}. The figure demonstrates how the RAG agent responds to a knowledge question helpful for diagnosis, along with the references used to generate the answer. As shown, the RAG agent’s responses are formulated strictly from the retrieved content. Based on this output, the final doctor agent produces a ranked list of potential diagnoses, each with a confidence score indicating the system’s certainty. 
% One key observation from example 2 is that integrating the retrieved information prompted a shift in the highest-ranked diagnosis, highlighting the system’s capacity to adapt to new findings.

% To evaluate our RADAR method’s efficacy, we compare it with three methods: a non agentic baseline consisting of a single LLM generating diagnosis, and two alternative multi-agent designs described in Section~\ref{ssec:agentic_baselines}. The Top-1 and Top-5 accuracy results across various LLMs are summarized in Table~\ref{table:top1acc} and Table~\ref{table:top5acc}.

% Overall, our approach consistently outperformed all other methods across models and metrics. For the Qwen3 model, our approach achieved a 7.97\% enhancement in Top-1 accuracy compared to the non-agent baseline. Furthermore, our RADAR configuration yielded a 10.19\% gain in Top-5 accuracy for the DeepSeek model. Notably, the improvements were more substantial when using open-source models compared to the closed-source models, indicating that our agentic coordination particularly benefits models with greater transparency and adaptability. 

\begin{table}[t]
\caption{Top 1 Accuracy comparison} \label{table:top1acc}
\begin{adjustbox}{width=8.5cm}
\begin{tabular}{lllll}
\hline
 Model & Single-agent & Collaboration & Challenge & RADAR  \\ \hline
 Gemini-2.0 & $45.55 \pm 1.58$ & $45.56 \pm 1.09$ & $41.48 \pm 1.0$9 & $\mathbf{48.54 \pm 0.29}$ \\ 
 GPT-4o &  $49.94 \pm 0.86$ & $48.61 \pm 1.11$ & $47.68 \pm 2.26$ & $\mathbf{54.40 \pm 1.02}$ \\
 Qwen3-32b &  $35.19 \pm 0.51$ & $36.85 \pm 2.11$ & $35.07 \pm 0.84$ & $\mathbf{43.16 \pm 3.16}$\\
 DeepSeek-R1-70B &  $35.47 \pm 2.72$ & $38.38 \pm 2.36$ & $37.89 \pm 0.37$ & $\mathbf{41.85 \pm 1.89}$ \\
 Medgemma-27B &  $31.00 \pm 1.20$ & $33.64 \pm 1.04$ & $36.06 \pm 1.05$ & $\mathbf{ 38.23 \pm 1.43}$ \\
\bottomrule
\end{tabular}
\end{adjustbox}

% \vspace{1em}  % NO NO 
\caption{Top 5 Accuracy comparison} \label{table:top5acc}
\begin{adjustbox}{width=8.5cm}
\begin{tabular}{lllll}
\hline
 Model & Single-agent & Collaboration & Challenge & RADAR  \\ \hline
 Gemini-2.0 & $66.21 \pm 1.40$ & $65.09 \pm 0.56$ & $68.98 \pm 1.22$ & $\mathbf{72.10 \pm 2.34}$ \\ 
 GPT-4o &  $68.10 \pm 1.65$ & $68.74 \pm 0.55$ & $69.29 \pm 1.17$ & $\mathbf{75.05 \pm 2.19}$ \\
 Qwen3-32b &  $52.54 \pm 0.39$ & $54.35 \pm 1.23$ & $55.01 \pm 2.45$ & $\mathbf{62.51 \pm 2.71}$ \\
 DeepSeek-R1-70B &  $54.62 \pm 2.53$ & $60.18 \pm 2.02$ & $59.11 \pm 1.85$ & $\mathbf{64.81 \pm 1.52}$ \\
 Medgemma-27B &  $56.21 \pm 0.54$ & $57.65 \pm 1.36$ & $61.53 \pm 0.78$ & $\mathbf{64.40 \pm 2.11}$ \\
 \bottomrule
\end{tabular}
\end{adjustbox}
\end{table}

% In contrast, the collaboration and challenger setup showed only slight enhancements compared to the non-agentic baseline, and, in certain instances, exhibitied a decline in performance. These results suggest that introducing multiple agents without the adequate domain-specific medical knowledge could create ambiguity. Moreover, considering that our dataset primarily consists of rare diseases, it is highly probable that incorporating additional specialized knowledge is critical for achieving better performance.

% Our method also significantly outperformed the results reported in the original NOVA dataset paper, which evaluated performance using a single LLM output. In comparison to their best Top-1 and Top-5 accuracies of 24.2\% and 38.4\% \cite{bercea2025nova}, respectively, our approach achieved substantially superior results. Among all configurations, the optimal performance was observed with GPT-4o model with our RADAR method, achieving a Top-1 accuracy of 54.40\% and Top-5 accuracy of 75.05\%, highlighting the strong synergy between advanced LLMs and medical knowledge injection by RAG. These results indicate that multi-agent coordination combined with retrieval mechanisms improves diagnostic accuracy and also scales effectively across diverse models, with the greatest relative gains observed in open-source models. 

\section{Conclusion}

We introduced \textbf{RADAR}, a retrieval-augmented, agentic framework for rare disease diagnosis in brain MRI.  
By coupling large language models with external medical knowledge, RADAR enhances diagnostic reasoning and provides interpretable, evidence-grounded outputs.  
Our results demonstrate that integrating retrieval mechanisms consistently improves diagnostic accuracy—particularly for open-source models—showing that explicit knowledge injection can complement model size.  Although the current system relies on radiologist-provided captions rather than direct image interpretation, bridging this gap between text-based reasoning and image-based understanding represents a key direction for future research.

\newpage
\section{Acknowledgments}
This project is supported by the Innovative Health Initiative Joint Undertaking (IHI JU) under grant agreement No 101132356 as part of the project PREDICTOM. 

\noindent PREDICTOM is supported by the Innovative Health Initiative Joint Undertaking (IHI JU), under Grant Agreement No 101132356. JU receives support from the European Union’s Horizon Europe research and innovation programme, COCIR, EFPIA, EuropaBio, MedTechEurope and Vaccines Europe. The UK participants are supported by UKRI Grant No 10083467 (National Institute for Health and Care Excellence), Grant No 10083181 (King's College London), and Grant No 10091560 (University of Exeter). University of Geneva is supported by the Swiss State Secretariat for Education, Research and Innovation Ref No 113152304. See www.ihi.europa.eu for more details.”

\noindent This work is supported by the DAAD programme under Konrad Zuse Schools of Excellence for Reliable AI (RelAI) 

\noindent C.I.B. is funded via the EVUK program (”Next-generation AI for Integrated Diagnostics”) of the Free State of Bavaria.

%  Examples of
% appropriate statements include:
% \begin{itemize}
%   \item ``No funding was received for conducting this study. The
%     authors have no relevant financial or non-financial interests to
%     disclose.'' 
%   \item ``This work was supported by […] (Grant numbers) and
%     […]. Author X has served on advisory boards for Company Y.'' 
%   \item ``Author X is partially funded by Y. Author Z is a Founder and
%     Director for Company C.''
% \end{itemize}

% References should be produced using the bibtex program from suitable
% BiBTeX files (here: strings, refs, manuals). The IEEEbib.bst bibliography
% style file from IEEE produces unsorted bibliography list.
% ------------------------------------------------------------------------- 
\bibliographystyle{IEEEbib}
\bibliography{strings,refs}

\end{document}